\documentclass[conference,letterpaper]{IEEEtran}
\usepackage[letterpaper, left=0.73in, right=0.73in, bottom=0.81in, top=0.86in]{geometry}
\IEEEoverridecommandlockouts   
\usepackage{balance} 
\usepackage{pgfplots}
\usepackage{relsize}
\pgfplotsset{compat=newest}
\usepackage{textcomp, gensymb}
\usepackage{textgreek}
\usepackage[symbol]{footmisc}

\usepackage[firstpageonly=true]{draftwatermark}

\usepackage{cancel}

\usepackage{amssymb}  
\usepackage{mathtools}

\usepackage{import}
\usepackage{pgfplots}
\usepackage{color}
\usepackage{tabularx} 
\usepackage{multirow}
\usepackage{pgf}
\usepackage{blindtext}
\usepackage{siunitx}
\usepackage{balance}
\usepackage{epsfig}
\usepackage{comment}
\usepackage{gensymb}

\renewcommand{\baselinestretch}{0.988}

\title{\LARGE \bf
Enhancing Robustness in Manipulability Assessment: \\ The Pseudo-Ellipsoid Approach
\thanks{This paper has been accepted for presentation at the IEEE/RSJ International Conference on Intelligent Robots and Systems (IROS), 2024.}
}

\author{ Erfan Shahriari$^{1,2}$, Kim Kristin Peper$^{1}$, Matej Hoffmann$^{3}$ and Sami Haddadin$^{1}$

\thanks{$^{1}$ Chair of Robotics and Systems Intelligence,
Munich Institute of Robotics and Machine Intelligence, Technical
University of Munich (TUM), Germany.
        {\tt\small firstname.lastname@tum.de}
        }%
\thanks{$^{2}$ Newman Laboratory for Biomechanics and
Human Rehabilitation, Department of Mechanical Engineering, Massachusetts Institute of Technology (MIT), USA.
        {\tt\small erfan@mit.edu}
        }%
\thanks{$^{3}$ Department of Cybernetics, Faculty of Electrical Engineering, Czech Technical University in Prague (CTU), Czech Republic.
        {\tt\small matej.hoffmann@fel.cvut.cz}
        }%
}

\begin{document}

\SetWatermarkAngle{0}
\SetWatermarkColor{black}
\SetWatermarkLightness{0.5}
\SetWatermarkFontSize{9pt}
\SetWatermarkVerCenter{30pt}
\SetWatermarkText{\parbox{30cm}{%
\centering This is the authors' final version of the manuscript published as:\\
\centering Shahriari, E.; Peper, K.K.; Hoffmann, M. \& Haddadin, S. (2024),\\
\centering Enhancing Robustness in Manipulability Assessment: The Pseudo-Ellipsoid Approach  \\
\centering  in '2024 IEEE/RSJ International Conference on Intelligent Robots and Systems (IROS)', pp. 1329-1336. (C) IEEE \\
}}

\maketitle
\thispagestyle{empty}

\begin{abstract}
Manipulability analysis is a methodology employed to assess the capacity of an articulated system, at a specific configuration, to produce motion or exert force in diverse directions. The conventional method entails generating a virtual ellipsoid using the system's configuration and model. Yet, this approach poses challenges when applied to systems such as the human body, where direct access to such information is limited, necessitating reliance on estimations. Any inaccuracies in these estimations can distort the ellipsoid's configuration, potentially compromising the accuracy of the manipulability assessment. To address this issue, this article extends the standard approach by introducing the concept of the manipulability \emph{pseudo-ellipsoid}. Through a series of theoretical analyses, simulations, and experiments, the article demonstrates that the proposed method exhibits reduced sensitivity to noise in sensory information, consequently enhancing the robustness of the approach.
\end{abstract}

\section{Introduction}

Ensuring the effectiveness of any system involves confirming that tasks align with its capabilities and operational requirements. For instance, when dealing with an articulated body, such as the human arm or a robot manipulator, one factor to consider is whether the arm's configuration is appropriate for the task at hand. The task may involve specific movements of the endpoint (e.g., the hand) or the application of force in a particular direction. Exploring these considerations in literature falls under the concept known as \emph{manipulability}. In essence, the study of manipulability delves into the effectiveness of mapping between different spaces---in our examples, the joint space and task space. By addressing these inquiries, one can assess the feasibility of executing various manipulation tasks in different configurations.

\begin{figure}
    \centering
    \includegraphics[width=\columnwidth]{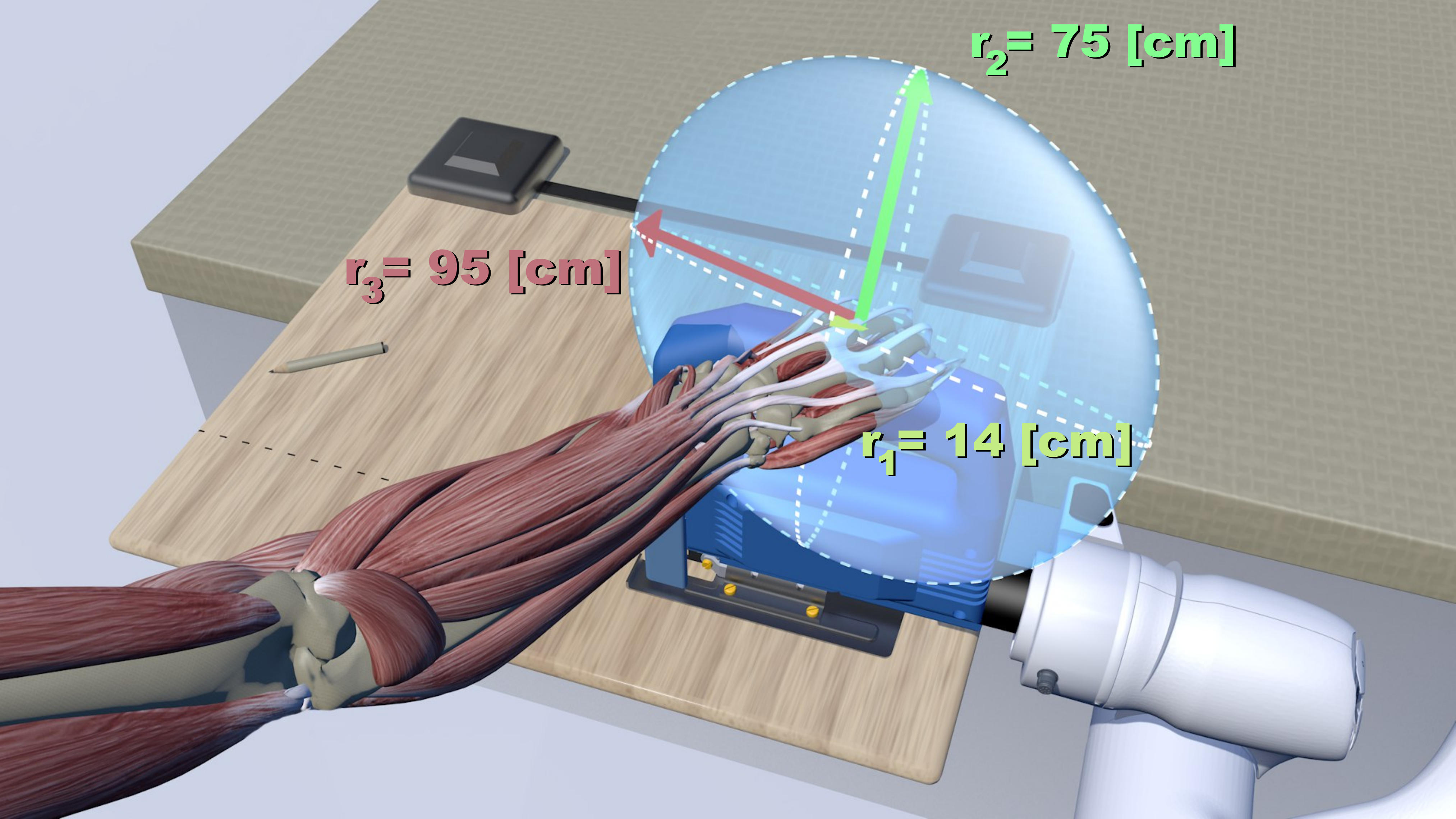}
    \caption{Downscaled visualization of the velocity manipulability ellipsoid and the principal radii for a right arm 12-DoF kinematics model, based on the VHP dataset \cite{Garner1999} with the following joint values (in radian): sternoclavicular [-0.31, 0.33], acromioclavicular [0.47, 0.1, -0.02], glenohumeral [0.52, 0.37, -0.98], humeroulnar [0.17], radioulnar [1.86] and radiocarpal [-0.11, -0.25].}
    \label{fig:skeleton_side}
\end{figure}

Introduced in \cite{Hanafusa1981} and further refined in \cite{Yoshikawa1985}, the concept of manipulability has served as a valuable tool for evaluating the effectiveness of force/motion mapping across different spaces of a manipulator. This concept has subsequently been applied to increasingly sophisticated systems, including dual-arm setups in \cite{Lee1989} and parallel manipulators in \cite{Merlet2006}. Among the pioneering works that extended this concept to the human body, the work by Lenar\v{c}i\v{c} et al. \cite{Lenarcic1991} stands out. In this context, the manipulability metric was introduced in \cite{Tanaka2004} and subsequently used to assess human force exertion ability in \cite{Tanaka2005}. Their later research in \cite{Tanaka2014} demonstrated a substantial correlation between the spatial characteristics of force manipulability and the concept of human operational comfort. Another study by Artemiadis et al. \cite{Artemiadis2011} revealed a high correlation between human manipulability and muscle activation as determined by EMG signals.

The representation of manipulability as an ellipsoid has found increasing application in human-centered robotics (see Fig.~\ref{fig:skeleton_side}). In \cite{Peternel2016} and \cite{Peternel2017}, the robot task frame was aligned based on the direction of the human force manipulability ellipsoid. Similarly, in \cite{Gopinathan2018}, the robot motion controller gains were adjusted based on the configuration of the manipulability ellipsoid. An assistive robot control scheme, based on manipulability, designed to aid humans in applying forces in challenging directions was presented in \cite{Petric2016}. This approach was later extended to consider human muscular force manipulability in \cite{Babic2017} and \cite{Petric2019}, incorporating considerations of muscular space ergonomics. In \cite{Wansoo2019}, the volume of the manipulability ellipsoid was utilized as an assessment metric to ergonomically optimize scenarios involving multi-human, multi-human collaboration. Finally, in \cite{lachner2020influence}, the limitations of conventional manipulability analysis approaches are addressed, particularly in terms of integrating different coordinates. This was tackled by considering both kinematic and inertial parameters in the derivation of the manipulability ellipsoid, similar to approaches used in \cite{hogan1984impedance} and \cite{khatib1995inertial}.

During the initial two decades following its formulation \cite{Hanafusa1981}, manipulability analysis was primarily applied to mechanical arms. It was not until the early 2000s that it was first utilized in human studies \cite{Tanaka2004}. An important difference  between a mechanical arm and a human limb from an engineering perspective lies in the fact that the joint configuration of a human limb cannot be directly accessed and needs to be estimated. Rapid progress in computer vision makes it possible to estimate the positions of human body keypoints from images or videos only (e.g., \cite{Cao2019_Openpose,xu2023vitpose++} for 2D estimation and \cite{goel_humans_2023,sun2023trace} for 3D estimation). However, this estimation is not free of errors. Moreover, joint angles need to be estimated in a second step. Yet, the assessment of manipulability heavily relies on the estimation of the joint configuration.  This challenge becomes particularly pronounced when the body configuration is such that small deviations in sensory data can have a dramatic impact on manipulability assessment.

In this article, building upon the concept of manipulability analysis, an approach is presented that is less sensitive to instantaneous changes in configuration data. The proposed approach replaces the conventional ellipsoid with the manipulability \emph{pseudo-ellipsoid}. This method proves particularly valuable for evaluating the manipulability of the human body, especially when relying on compact but inaccurate body configuration estimations, such as those from RGB camera input alone. In the following sections, both the conventional ellipsoid and the novel pseudo-ellipsoid concepts are introduced. Subsequently, a series of theoretical, numerical, and real experiments are conducted to compare the effectiveness of both approaches in the presence of sensory inaccuracies.

\section{Fundamentals: Manipulability Ellipsoid}
Manipulability analysis is a method used to evaluate the manipulation capabilities of robots, specifically their ability to execute movements or exert wrenches in various directions. Several methodologies have been introduced for conducting such analyses. Nonetheless, a common visualization for these methodologies emerges in the form of a pair of dual ellipsoids. These ellipsoids are fundamentally linked through their core matrices, which are inverses of one another, serving distinct yet complementary purposes: one delineates the robot's movement capabilities, while the other illustrates its potential for applying wrenches in all directions.

Consider a robot with $n$ degrees of freedom whose configuration vector is represented as $\boldsymbol{q}~\in~\mathbb{R}^n$. Moreover, let $\boldsymbol{J}(\boldsymbol{q})~\in~\mathbb{R}^{m \times n}$ denote its Jacobian matrix, such that
\begin{align}
\dot{\boldsymbol{x}} = \boldsymbol{J}(\boldsymbol{q}) \dot{\boldsymbol{q}},
\end{align}
where $\dot{\boldsymbol{x}} \in \mathbb{R}^m$ is the end-point's twist in an $m$-dimensional Cartesian space. The motion-based manipulability ellipsoid for this robot can be constructed using a symmetric positive definite core matrix $\boldsymbol{\Lambda}(\boldsymbol{q})~\in~\mathbb{R}^{m \times m}$ derived as 
\begin{align}
    \boldsymbol{\Lambda}(\boldsymbol{q}) = \boldsymbol{J}(\boldsymbol{q}) \boldsymbol{\Upsilon} \boldsymbol{J}^T(\boldsymbol{q}). \label{eq:core}
\end{align}
Here, $\boldsymbol{\Upsilon} \in \mathbb{R}^{n \times n}$ may assume various forms depending on the chosen approach. For instance, when it is set as an identity matrix, \eqref{eq:core} denotes the core matrix of the widely recognized velocity manipulability ellipsoid \cite{Yoshikawa1985}. This ellipsoid determines the attainability of end-point twists in various directions, achieved by selecting configuration velocity vectors of unit norm, i.e., $\| \dot{\boldsymbol{q}} \| = 1$. Orthogonal to this ellipsoid is its dual counterpart, known as the force manipulability ellipsoid, whose core matrix is the inverse of that for the velocity manipulability ellipsoid. The force manipulability ellipsoid illustrates the feasibility of applying wrenches in various directions for a unit-norm configuration-space forces/torques.

Considering $\boldsymbol{M}(\boldsymbol{q}) \in \mathbb{R}^{n \times n}$ as the mass matrix of the robot and defining $\boldsymbol{\Upsilon} = \boldsymbol{M}^{-1}(\boldsymbol{q})$, \eqref{eq:core} characterizes the mobility end-point tensor \cite{hogan1984impedance}. This tensor governs the range of achievable end-point motions in different directions when unit-norm wrench vectors are applied to the end-point, while the body is at rest. This concept is similarly employed in other studies, such as in \cite{khatib1995inertial}, which introduces the concept of the belted ellipsoid to represent inertial properties in the task space, and in \cite{khatib2004whole}, which distinguishes between inertial properties associated with individual tasks a robot is supposed to perform. As elaborated in \cite{lachner2020influence}, integrating robot dynamics into manipulability analysis can alleviate certain limitations inherent in conventional approaches, such as their susceptibility to the influence of coordinate choice or their challenges in handling kinematically redundant robots.

Irrespective of the chosen $\boldsymbol{\Upsilon}$, a motion-based manipulability ellipsoid, centered at the end-point $\boldsymbol{x}$, with its core matrix $\boldsymbol{\Lambda}(\boldsymbol{q})$ derived in \eqref{eq:core}, can be visualized in the Cartesian space using $\boldsymbol{x}_\mathrm{ell} \in \mathbb{R}^m$ satisfying
\begin{align}
(\boldsymbol{x}_\mathrm{ell}-\boldsymbol{x})^T \boldsymbol{\Lambda}^{-1}(\boldsymbol{q}) (\boldsymbol{x}_\mathrm{ell}-\boldsymbol{x}) = 1. \label{eq:mani}
\end{align}
Considering $n_\Lambda \leq m$ eigenvalues for the matrix $\boldsymbol{\Lambda}(\boldsymbol{q})$ with values $\lambda_i \in \mathbb{R}_{> 0}$, the ellipsoid \eqref{eq:mani} would have $n_\Lambda$ principal radii\footnote{Also known as principal semi-axes.} with lengths $r_i = \sqrt{\lambda_i}$ along the eigenvectors $\boldsymbol{\nu}_i \in \mathbb{R}^m$ of the matrix $\boldsymbol{\Lambda}(\boldsymbol{q})$. The length of each ellipsoid's principal radius indicates the significance of the manipulability feature along the direction of it. For example, in the case of a velocity manipulability ellipsoid (i.e., when $\boldsymbol{\Upsilon}$ is an identity matrix), these lengths demonstrate the feasibility of end-point motions along their respective directions for unit-norm configuration velocity vectors -- see Fig.~\ref{fig:skeleton_side}.

To assess manipulability along a specific direction (unit vector) $\boldsymbol{\nu} \in \mathbb{R}^m$, one needs to determine the ellipsoid's radius length $r \in \mathbb{R}_{\geq 0}$ corresponding to this direction. For that, the following condition can be considered \cite{kreyszig1967advanced}.
\begin{align}
    \mathlarger{\sum}\limits_{i \mid r_i \neq 0} \frac{(r \cos(\theta_i))^2}{r_i^2} = 1. \label{eq:ellEq1}
\end{align}
Here, $\theta_i$ is the angle between the unit vector $\boldsymbol{\nu}$ and the eigenvectors $\boldsymbol{\nu}_i$ indicating the directions of the ellipsoid's principal axes such that
\begin{align}
    \cos(\theta_i) = \boldsymbol{\nu}_i^T {\boldsymbol{\nu}}. \label{eq:ellCos}
\end{align}
Employing \eqref{eq:ellCos} in \eqref{eq:ellEq1} leads to
\begin{align}
    r^2 \mathlarger{\sum}\limits_{i \mid r_i \neq 0} \frac{(\boldsymbol{\nu}_i^T {\boldsymbol{\nu}})^2}{r_i^2} = 1,
\end{align}
from which $r$ can be expressed:
\begin{align}
    r = \frac{1}{\sqrt{\mathlarger{\sum}_{i \mid r_i \neq 0} \frac{(\boldsymbol{\nu}_i^T \boldsymbol{\nu})^2}{r_i^2}}}. \label{eq:R}
\end{align}
Note that for $r_i = 0$, the ellipsoid loses a dimension along the direction of $\boldsymbol{\nu}_i$.  In this case, only if the vector ${\boldsymbol{\nu}}$ is orthogonal to $\boldsymbol{\nu}_i$, \eqref{eq:R} can be employed. Otherwise, $r$ would be zero.

The ellipsoid's radius $r$, derived in \eqref{eq:R}, acts as a metric for evaluating the manipulability of an articulated body along the direction $\boldsymbol{\nu}$. The accuracy of this assessment depends on the precision of constructing the ellipsoid. Essentially, this precision correlates with the accuracy of the components used for deriving the ellipsoid's core matrix. Considering \eqref{eq:core}, these components encompass the configuration vector $\boldsymbol{q}$, the kinematic model utilized in the Jacobian matrix $\boldsymbol{J}(\boldsymbol{q})$, and the dynamics model, which may be incorporated into $\boldsymbol{\Upsilon}$. When these components are reliably determined, $r$ can serve as a dependable metric. Otherwise, an alternative metric derived from manipulability analysis may be employed, which is less sensitive to the accuracy of these values. The subsequent section aims to explore this alternative approach.

\begin{figure}
    \centering
    \includegraphics[width=\columnwidth]{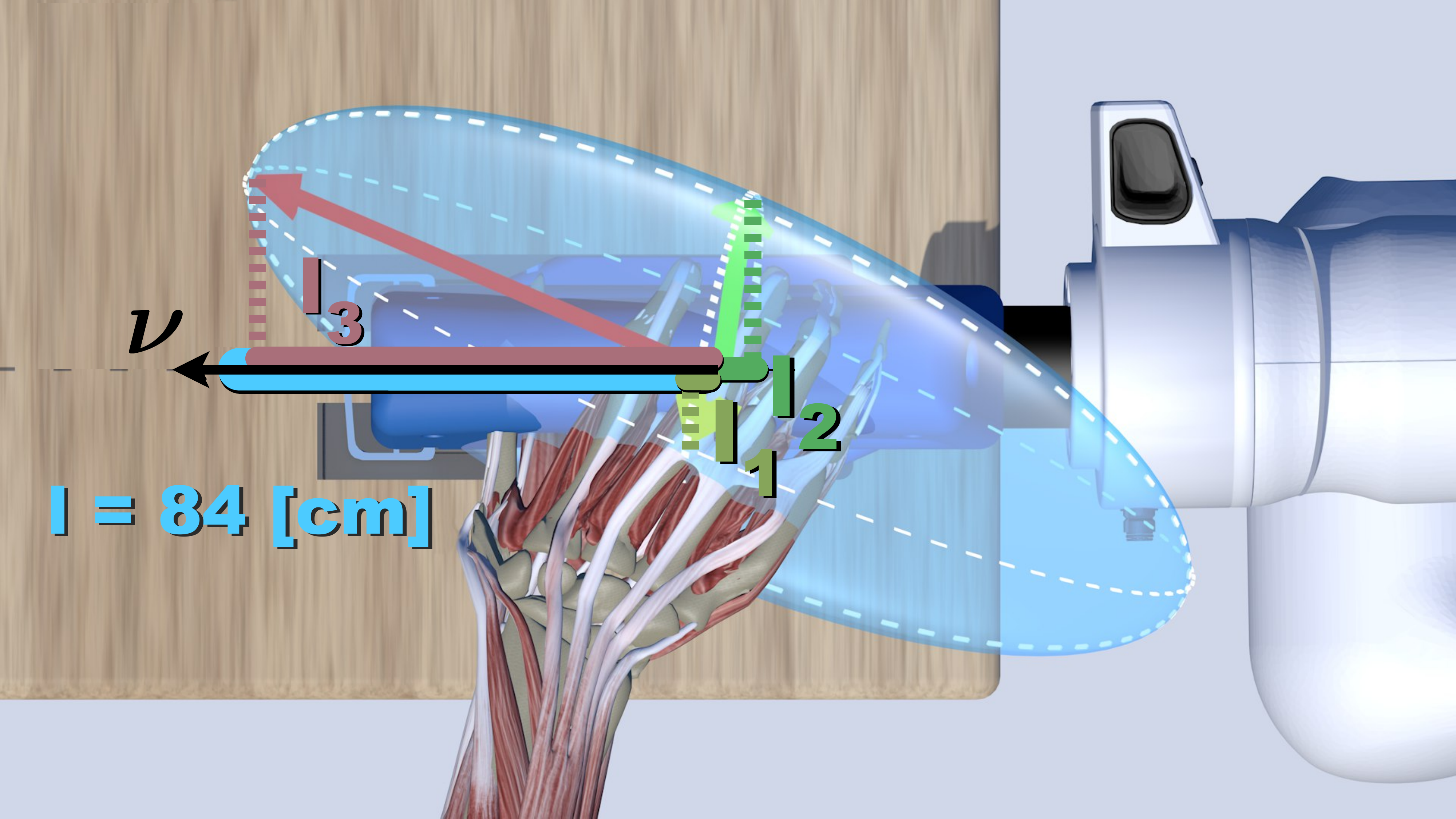}
    \caption{Ellipsoid's principal radii in Fig.~\ref{fig:skeleton_side} projected onto the direction $\boldsymbol{\nu}$, and the resulting projection norm $l$.}
    \label{fig:skeleton_top}
\end{figure}

\section{Manipulability Pseudo-Ellipsoid}

As previously discussed, the significance of the manipulability feature along the unit vector $\boldsymbol{\nu}$ can be determined by the length of the principal radius $r_i$ when $\boldsymbol{\nu}$ is aligned with the eigenvector $\boldsymbol{\nu}_i$, i.e., when\footnote{Note that in this case, for other eigenvectors $\boldsymbol{\nu}_j$, $\boldsymbol{\nu}_j^T \boldsymbol{\nu} = 0$.} $\boldsymbol{\nu}_i^T \boldsymbol{\nu} = 1$. Following this principle, an alternative approach for determining the manipulability along $\boldsymbol{\nu}$ could involve projecting each principal radius onto the direction $\boldsymbol{\nu}$, resulting in the projection length $l_i \in \mathbb{R}$, as 
\begin{align}
    l_i = | r_i \boldsymbol{\nu}_i^T \boldsymbol{\nu} |. \label{eq:li}
\end{align}
With $\boldsymbol{l}_\Lambda = [l_1, l_2, \hdots, l_{n_\Lambda}]$ representing the vector of all projection lengths, an alternative manipulability assessment metric along $\boldsymbol{\nu}$ can be defined as the norm $l$, derived as
\begin{align}
    l = \sqrt{\sum_{i=1}^{n_\Lambda} l_i^2}. \label{eq:l}
\end{align}
By integrating \eqref{eq:ellCos} into \eqref{eq:li}, \eqref{eq:l} transforms into
\begin{align}
    l = \sqrt{\sum_{i=1}^{n_\Lambda} r_i^2 \cos^2(\theta_i)}. \label{eq:lcos}
\end{align}
Given $r_\mathrm{min}$ and $r_\mathrm{max}$ as the minimum and maximum values of $r_i$, respectively, the following holds
\begin{align}
    \sqrt{\sum_{i=1}^{n_\Lambda} \hspace{-1mm} r_\mathrm{min}^2\hspace{-1mm}  \cos^2(\theta_i)} \hspace{-1mm} \leq \hspace{-1mm} \sqrt{\sum_{i=1}^{n_\Lambda} \hspace{-1mm} r_i^2 \cos^2(\theta_i)} \hspace{-1mm}  \leq \hspace{-1mm}   \sqrt{\sum_{i=1}^{n_\Lambda} \hspace{-1mm}  r_\mathrm{max}^2 \hspace{-1mm}  \cos^2(\theta_i)}.
\end{align}
As $\sqrt{\sum_{i=1}^{n_\Lambda} \cos^2(\theta_i)}$ represents the norm of a unit vector defined in a coordinate system aligned with the ellipsoid's principal axes, one can express:
\begin{align}
    r_\mathrm{min} \hspace{-1mm}  \underbrace{\sqrt{\sum_{i=1}^{n_\Lambda} \hspace{-1mm}  \cos^2(\theta_i)}}_{= 1} \hspace{-1mm}  \leq \hspace{-1mm}  \underbrace{\sqrt{\sum_{i=1}^{n_\Lambda} \hspace{-1mm}  r_i^2 \cos^2(\theta_i)}}_{ l} \hspace{-1mm}  \leq \hspace{-1mm}  r_\mathrm{max} \underbrace{\sqrt{\sum_{i=1}^{n_\Lambda} \hspace{-1mm}  \cos^2(\theta_i)}}_{= 1}, 
\end{align}    
which results in 
\begin{align}
        r_\mathrm{min} \leq l \leq r_\mathrm{max}.
\end{align}
Hence, the projection norm $l$ consistently falls between the longest and shortest radii lengths. A higher value of $l$ indicates a greater likelihood that the direction $\boldsymbol{\nu}$ aligns with the longest radius of the manipulability ellipsoid; see Fig.~\ref{fig:skeleton_top}.

\begin{figure}[t]
\centering
\includegraphics[width=1\columnwidth]{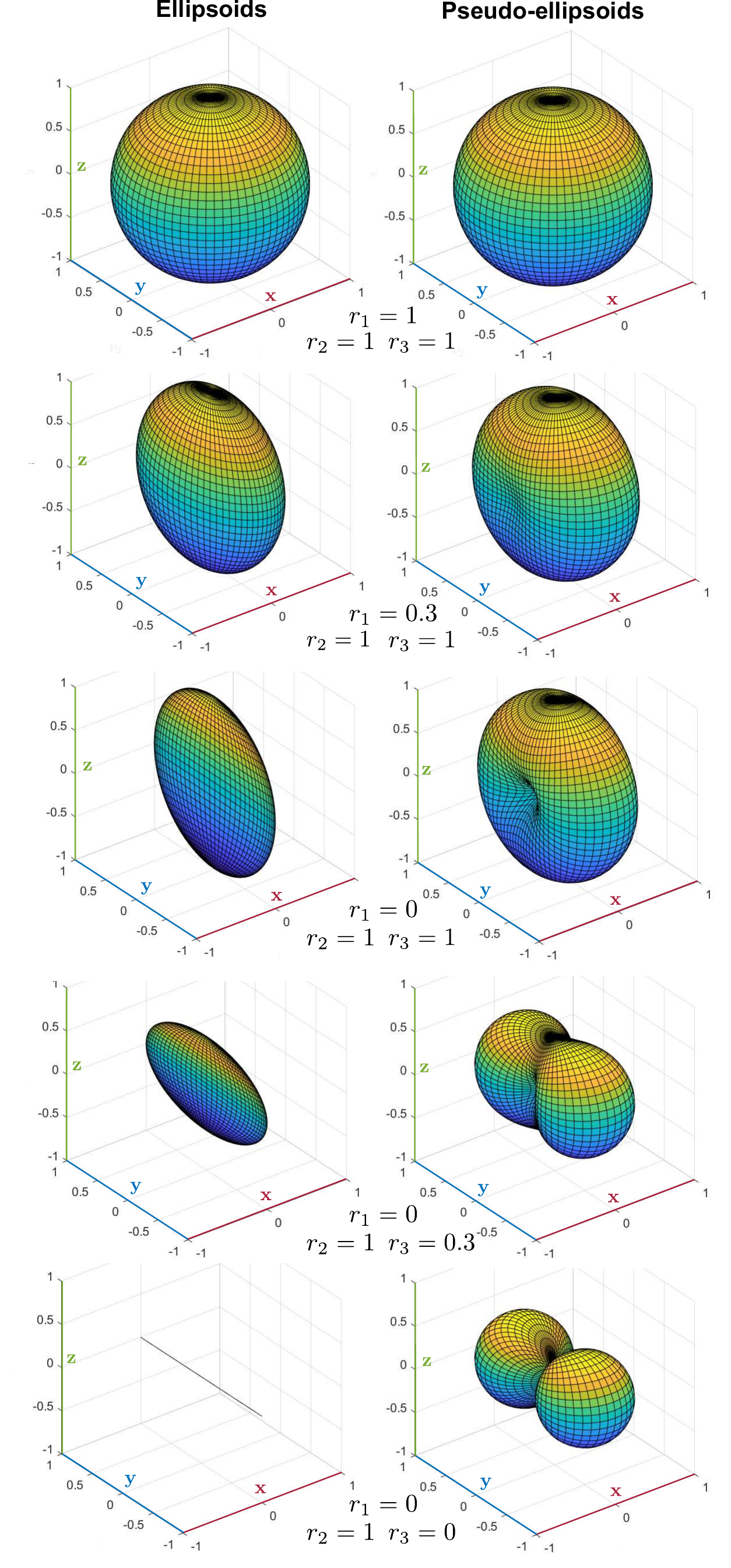}
\caption{Ellipsoids on the left, and their corresponding pseudo-ellipsoids on the right. The principal radii with the lengths $r_1$, $r_2$ and $r_3$ are along the axes $x$, $y$ and $z$, respectively.}
\label{fig:ellipsoids}
\end{figure}

Considering all unit vectors $\boldsymbol{\nu}$ and visualizing all potential radii with lengths $r$, as derived in \eqref{eq:R}, the manipulability ellipsoid is constructed with its principal axes aligned along $\boldsymbol{\nu}_i$. When this process is repeated using the norm $l$ instead of $r$, it yields a new geometric shape that is aligned with the original ellipsoid but has a smoother surface. We refer to this shape as the \textit{pseudo-ellipsoid}. The proposed method for evaluating manipulability along the direction $\boldsymbol{\nu}$ essentially comes down to determining the radius length of the manipulability pseudo-ellipsoid along $\boldsymbol{\nu}$. Figure \ref{fig:ellipsoids} provides visual examples of ellipsoids alongside their corresponding pseudo-ellipsoids.

In accordance with manipulability analysis principles, the radius length $r$ of the ellipsoid is identified as a more precise evaluation metric compared to the pseudo-ellipsoid's radius length $l$. Nonetheless, the computation of $l$, as demonstrated in \eqref{eq:l}, is simpler and circumvents the singularity issues (i.e., division by zero) that are encountered in the derivation of $r$, as detailed in \eqref{eq:R}. Moreover, employing $l$ rather than $r$ offers a significant advantage due to its lower sensitivity to inaccuracies. This is particularly crucial in scenarios where a precise model of the articulated body under analysis is lacking or if the sensors used for determining the configuration vector $\boldsymbol{q}$ are imprecise orprone to noise. A typical example is when applying manipulability analysis to a human limb, where, unlike for mechanical robots, there is no direct access to proprioceptive sensory information, and the configuration vector $\boldsymbol{q}$ can only be estimated. In this case, a less sensitive metric would be much more appropriate. In line with these concerns, the following section compares the sensitivity of the two proposed metrics to the configuration of the ellipsoid.

\section{Sensitivity Analysis}
The configuration of the manipulability ellipsoid is contingent upon the configuration vector $\boldsymbol{q}$. Consequently, any sensory inaccuracies in $\boldsymbol{q}$ can directly impact the estimated configuration of the ellipsoid, including the principal axes $\boldsymbol{\nu}_i$, radii lengths $r_i$, and ultimately the evaluation metrics $r$ and $l$ associated with the radius length of the conventional ellipsoid and the novel pseudo-ellipsoid, derived in \eqref{eq:R} and \eqref{eq:l}, respectively. In fact, considering \eqref{eq:ellEq1} and \eqref{eq:lcos}, \eqref{eq:R} and \eqref{eq:l} can be reformulated as
\begin{align}
    r = \frac{1}{\sqrt{U_r}}, \hspace{6mm}
    l = \sqrt{U_l},
\end{align}
where
\begin{align}
    U_r = \mathlarger{\sum}_{i=1}^{n_\Lambda}\frac{\cos^2(\theta_i)}{r_i^2}, \hspace{6mm}
    U_l = \mathlarger{\sum}_{i=1}^{n_\Lambda} r_i^2 \cos^2(\theta_i), \label{eq:UU}
\end{align}
and thus
\begin{align}
    \frac{\partial r}{\partial U_r} &= - \frac{1}{2 (\sqrt{U_r})^3} = - \frac{r^3}{2}, \\
    \frac{\partial l}{\partial U_l} &= \frac{1}{2 \sqrt{U_l}} = \frac{1}{2 l}.
\end{align}
Moreover, considering \eqref{eq:UU}, the following holds
\begin{align}
    \frac{\partial U_r}{\partial r_i} &= -\frac{2 \cos^2(\theta_i)}{r_i^3}, \\
    \frac{\partial U_r}{\partial \theta_i} &= -\frac{2 \cos{\theta_i} \sin{\theta_i}}{r_i^2} = -\frac{\sin(2\theta_i)}{r_i^2} \\
    \frac{\partial U_l}{\partial r_i} &= 2 r_i \cos^2(\theta_i) \\
    \frac{\partial U_l}{\partial \theta_i} &= - 2 r_i^2 \cos(\theta_i) \sin(\theta_i) = -r_i^2 \sin(2\theta_i). 
\end{align}
As a result the sensitivity of $r$ and $l$ with respect to $r_i$ and $\theta_i$ can be derived as
\begin{align}
    \frac{\partial r}{\partial r_i} &= \frac{\partial r}{\partial U_r} \frac{\partial U_r}{\partial r_i} = \frac{r^3 \cos^2(\theta_i)}{r_i^3}, \label{eq:r_ri} \\
    \frac{\partial l}{\partial r_i} &= \frac{\partial l}{\partial U_l} \frac{\partial U_l}{\partial r_i} = \frac{r_i \cos^2(\theta_i)}{l}, \label{eq:l_ri} \\ 
    \frac{\partial r}{\partial \theta_i} &= \frac{\partial r}{\partial U_r} \frac{\partial U_r}{\partial \theta_i} = \frac{r^3 \sin(2\theta_i)}{2 r_i^2}, \label{eq:r_thetai} \\
    \frac{\partial l}{\partial \theta_i} &= \frac{\partial l}{\partial U_l} \frac{\partial U_l}{\partial \theta_i} = -\frac{r_i^2\sin(2\theta_i)}{2l}. \label{eq:l_thetai}
\end{align}

According to \eqref{eq:r_ri} and \eqref{eq:r_thetai}, when a principal radius length $r_i$ approaches zero but the ellipsoid radius $r$ does not (i.e., $\boldsymbol{\nu}$ aligns with another long principal radius and $\theta_i \approx \frac{\pi}{2}$), the terms $\cos(\theta_i)$ and $\sin(2\theta_i)$ may have small values. However, ${\partial r}/{\partial r_i}$ and ${\partial r}/{\partial \theta_i}$ become large due to the higher power of $r_i$ in the denominators. This leads to a situation where $r$ becomes highly sensitive to potential changes in $\boldsymbol{q}$, which can result in variations of $r_i$ and $\boldsymbol{\nu}_i$. On the other hand, considering \eqref{eq:l_ri} and \eqref{eq:l_thetai}, although it is possible for $l$ to approach zero while $r_i$ does not (e.g., when $\boldsymbol{\nu}$ aligns with another short principal radius and $\theta_i \approx \frac{\pi}{2}$), the small numerator terms $\cos(\theta_i)$ and $\sin(2\theta_i)$ will have either higher or equal powers compared to the small $l$ in the denominators. Consequently, $l$ would not be as sensitive as $r$ to small changes in the estimated body configuration $\boldsymbol{q}$. Increased sensitivity implies that variations in the estimated ellipsoid's configuration due to sensory uncertainties (e.g., noise) may lead to abrupt changes in the assessment metric. Hence, while still employing the manipulability concept, we opt to use  the pseudo-ellipsoid's radius~$l$ instead of the ellipsoid's radius~$r$ to reduce the sensitivity of the assessment to sensory data.

\section{Experimental Validation}

\begin{figure}
\centering
\includegraphics[width=1\columnwidth]{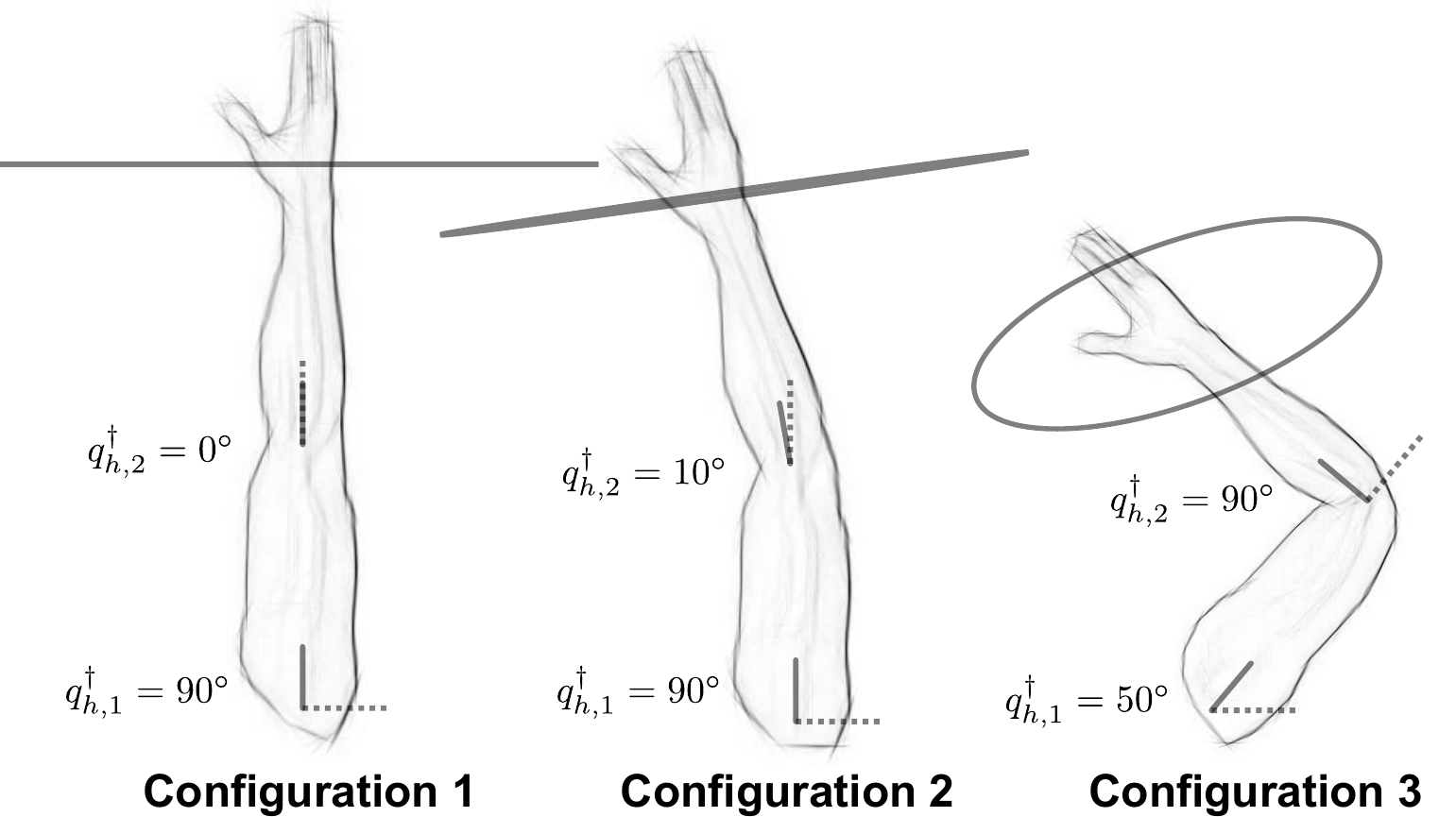}
\caption{Simulation scenario -- A 2-DoF arm kinematics model in three different configurations with their associated manipulability ellipses.}
\label{fig:threeArms}
\end{figure}

This section explores two scenarios to assess the effectiveness of two proposed manipulability analysis approaches: one utilizing the radius length $r$ of the conventional ellipsoid, and the other employing the radius length $l$ of the novel pseudo-ellipsoid, as derived in \eqref{eq:R} and \eqref{eq:l}, respectively. For the derivation of the ellipsoid's core matrix $\boldsymbol{\Lambda}(\boldsymbol{q})$ in \eqref{eq:core}, the matrix $\boldsymbol{\Upsilon}$ is set to an identity matrix, resulting in the so-called velocity manipulability ellipsoid. It is worth noting that this choice does not influence the comparisons. In the first scenario, a planar arm is simulated to illustrate the impact of configuration variations on the estimation of the manipulability metrics $r$ and $l$. This simplified choice highlights that even with a small number of variables involved, the impact can be significant. In the second scenario, a series of real experiments are conducted to evaluate the effectiveness of the two metrics when the human arm configuration is estimated using computer vision techniques only from an image.

\begin{figure}
\centering
\includegraphics[width=1\columnwidth]{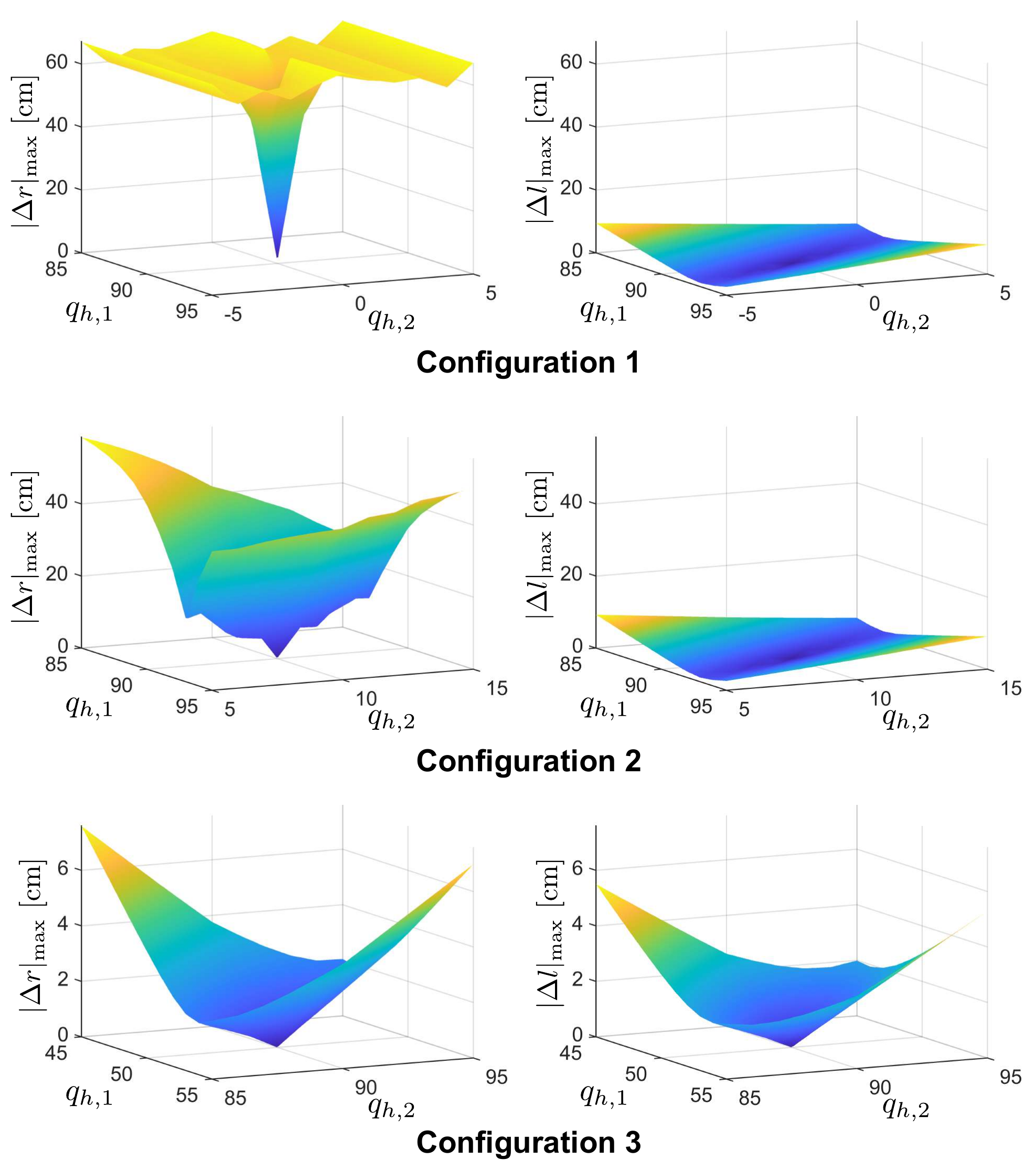}
\caption{Simulation results -- Maximum deviation of the estimated metrics: manipulability ellipsoid radius $r$ (left) and manipulability pseudo-ellipsoid radius $l$ (right) for changes in values $q_{h,1}$ and $q_{h,2}$ associated with different body configurations; see Fig. \ref{fig:threeArms}.}
\label{fig:maxDeviation}
\end{figure}

\subsection{Simulation -- Influence of Configuration Variation}
To better understand the impact of errors in configuration estimation on the assessment metrics $r$ and $l$, an example of a 2-DoF planar arm is examined with each link measuring $30$ cm in length, as depicted in Fig.~\ref{fig:threeArms}. For each configuration, there is an ideal manipulability ellipse\footnote{A 2-dimensional ellipsoid.} determined by the actual configuration values $q_{h,1}^\dag$ and $q_{h,2}^\dag$. Depending on the vector $\boldsymbol{\nu}$, the ideal metrics $r^\dag$ and $l^\dag$ can be calculated using \eqref{eq:R} and \eqref{eq:l}, respectively, considering the principal radii lengths and directions of the ideal ellipse. If the estimated configuration $\boldsymbol{q}_h$ deviates from the actual one $\boldsymbol{q}_h^\dag$, the estimated ellipse, and consequently the estimated metrics $r$ and $l$, differ from the ideal values due to the following errors.
\begin{align}
    \Delta r = r^\dag - r, \hspace{6mm}  \Delta l = l^\dag - l.
\end{align}
Considering all possible directions $\boldsymbol{\nu}$ and thus all potential values for $\Delta r$ and $\Delta l$, Fig.~\ref{fig:maxDeviation} illustrates the maximum absolute errors (i.e., $|\Delta r|_\mathrm{max}$ and $|\Delta l|_\mathrm{max}$) for all combinations of estimated values $q_{h,1}$ and $q_{h,2}$ within a range of $\pm 5$ degrees from the actual values $q_{h,1}^\dag$ and $q_{h,2}^\dag$. It is evident that when the arm is close to a singularity, even a slight estimation imprecision in one joint angle can lead to significant errors in $r$. Conversely, this is not the case for $l$, and consequently, even in the presence of considerable noise in the configuration vector estimation during manipulability evaluation, the assessment metric $l$ would not undergo significant changes. This becomes particularly crucial when the manipulability assessment metric is employed to execute an action, such as adjusting a control parameter. Sudden and drastic changes in this metric could lead to undesirable behavior or inconvenience. It should be mentioned that much like the radius $r$, the metric $l$ is derived from the arm's manipulability ellipse. However, it serves as a less stringent (and therefore more robust) tool for assessing manipulability, making it more suitable in the presence of noisy body configuration estimation.

\subsection{Experiment -- Effect of Noisy Configuration Estimation}
As previously pointed out, the velocity manipulability ellipsoid---utilized in this section by setting $\boldsymbol{\Upsilon}$ in \eqref{eq:core} to be an identity matrix---visualizes the feasibility of end-point motions in all directions for a unit-norm configuration velocity vector \cite{Yoshikawa1985}. In this context, an experiment is devised to evaluate the efficacy of the manipulability pseudo-ellipsoid relative to the conventional ellipsoid approach in situations where the estimation of the configuration vector is prone to error. Specifically, the manipulability assessment is carried out on a human arm whose configuration is estimated using computer vision techniques only from an RGB-D camera image. A motion capture system serves as a reference.

\begin{figure}
\centering
\includegraphics[width=0.9\columnwidth]{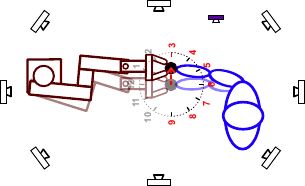}
\caption{Real Experiment scenario (configuration 1) -- Subjects were asked to perform instantaneous short motions along seven directions according to an imaginary horizontal clock face. The RGB-D camera (black icon) and the motion capture system cameras are depicted.}
\label{fig:viconSchem}
\end{figure}

\begin{figure}
\centering
\includegraphics[width=0.8\columnwidth]{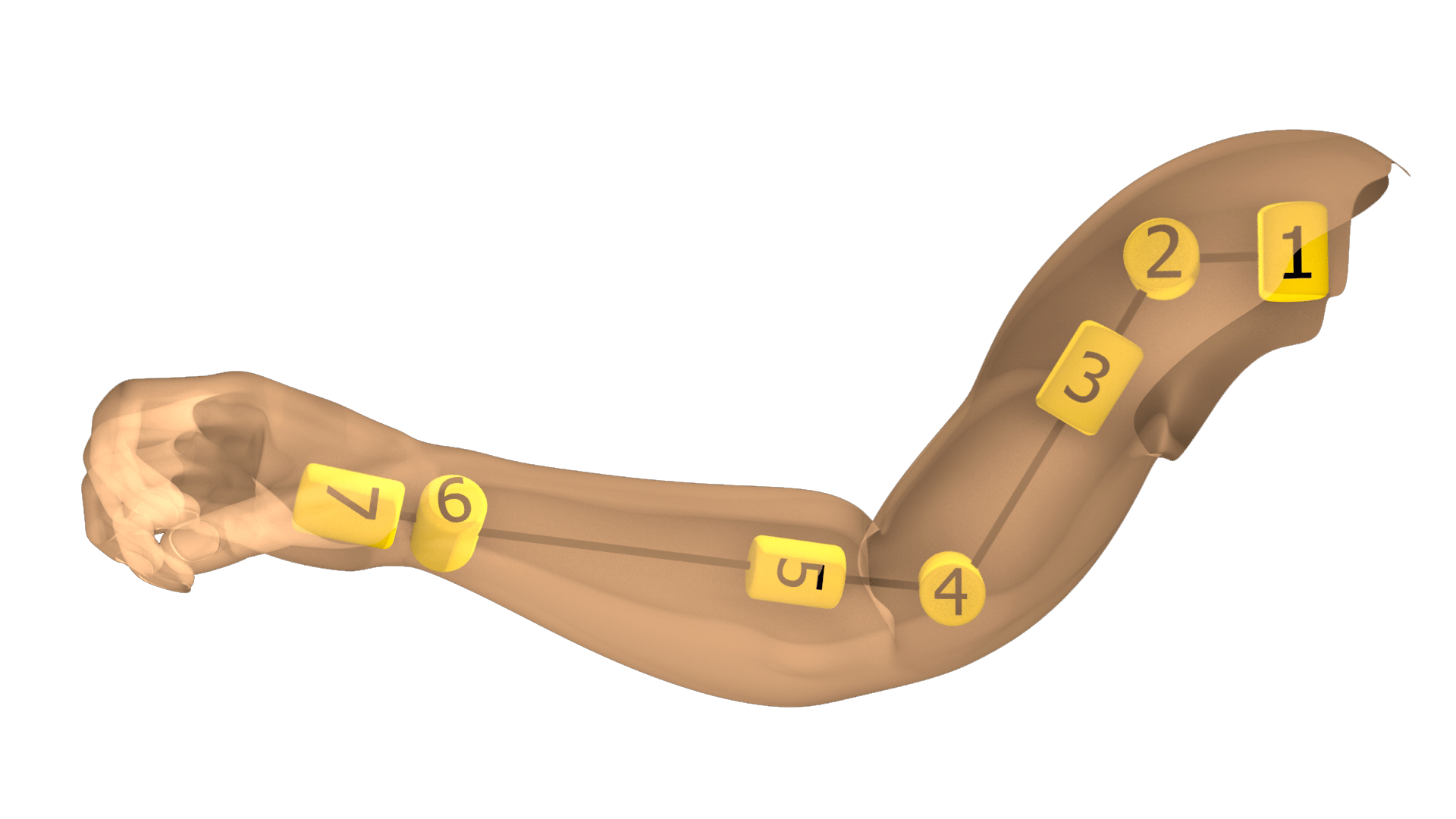}
\caption{Arm kinematic structure introduced in \cite{Lenarcic1994}. During the experiments, only the 1st, 2nd, and 4th  joints were considered to obtain the Jacobian matrix.}
\label{fig:kin}
\end{figure}

\begin{figure}
\centering
\includegraphics[width=0.95\columnwidth]{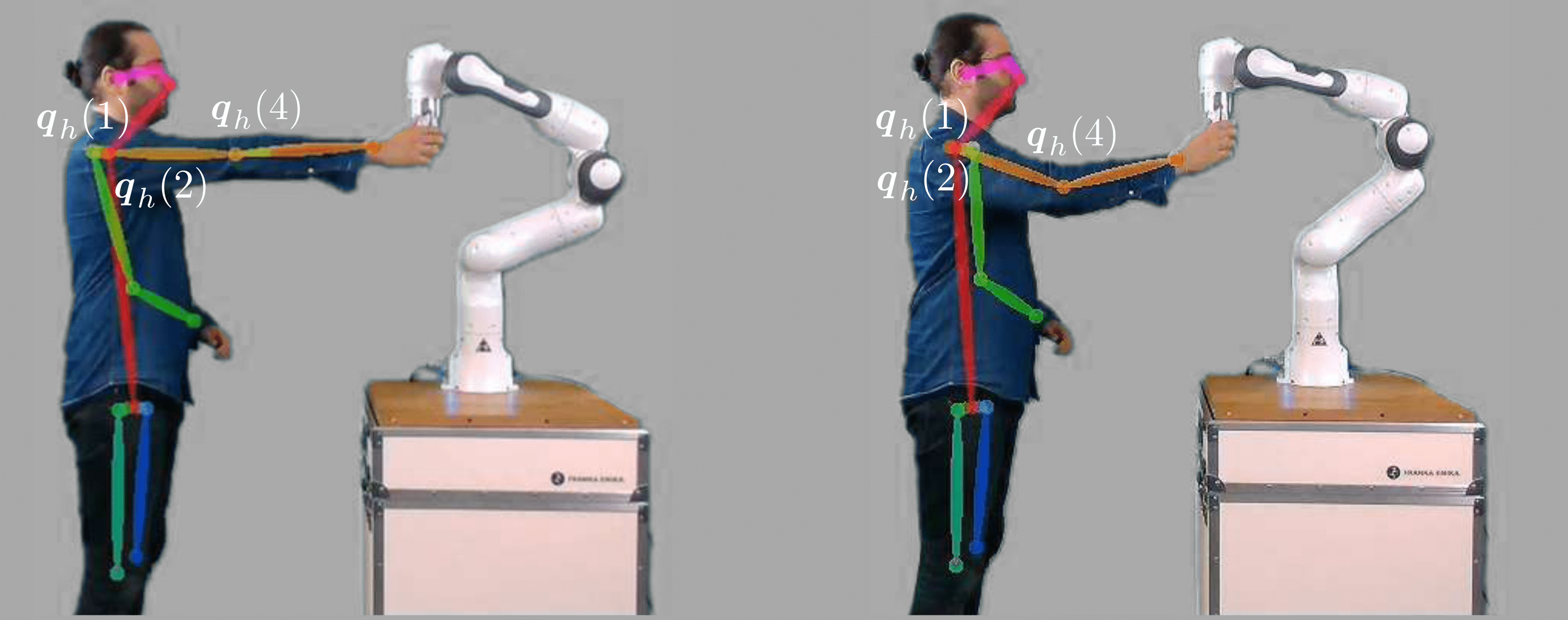}
\caption{Real Experiment setup -- Noisy configuration estimation. The locations of the body keypoints were estimated from an RGB-D camera image using the OpenPose library. Subsequently, the body joint angles were estimated using the reduced kinematics model depicted in Fig.~\ref{fig:kin}.}
\label{fig:openPose}
\end{figure}

Three participants were instructed to move their hand along seven distinct horizontal Cartesian directions, starting from two different initial arm configurations, as depicted in Fig.~\ref{fig:viconSchem}. The participants are identified as Subject 1, a male with a height of $176$ cm, Subject 2, a female with a height of $163$ cm, and Subject 3, a male with a height of $189$ cm. In order to capture the body configuration for deriving the manipulability assessment metrics, participants were recorded with an  RGB-D camera (Intel RealSense D435) and positions of 25 body keypoints estimated using computer vision algorithms---OpenCV3 \cite{Bradski2018} and in particular  OpenPose \cite{Cao2018, Cao2019} -- see Fig.~\ref{fig:openPose}. Subsequently, the keypoint positions were deprojected using the aligned depth image. These 3D keypoint positions were then inputted into a simplified kinematics model of the human arm, as illustrated in Fig.~\ref{fig:kin}, to calculate the human joint angles $\boldsymbol{q}_h$ and the human Jacobian matrix $\boldsymbol{J}_h(\boldsymbol{q}_h)$. Since the BODY-25 model considers only three keypoints for each arm (the wrist, elbow, and shoulder), the positions of these keypoints were utilized to determine the values of only three\footnote{More sophisticated approaches could be used to estimate other joints.} joints associated with arm abduction/adduction, arm flexion/extension, and elbow flexion/extension. These are referred to as joint 1, 2, and 4, respectively, in Fig.~\ref{fig:kin}. As a result, only three joints of the right arm kinematics were considered when constructing the body Jacobian matrix $\boldsymbol{J}_h(\boldsymbol{q}_h)$ using the simplified kinematics model introduced in \cite{Lenarcic1994}, as illustrated in Fig.~\ref{fig:kin}.

\begin{figure}
\centering
\includegraphics[width=\columnwidth]{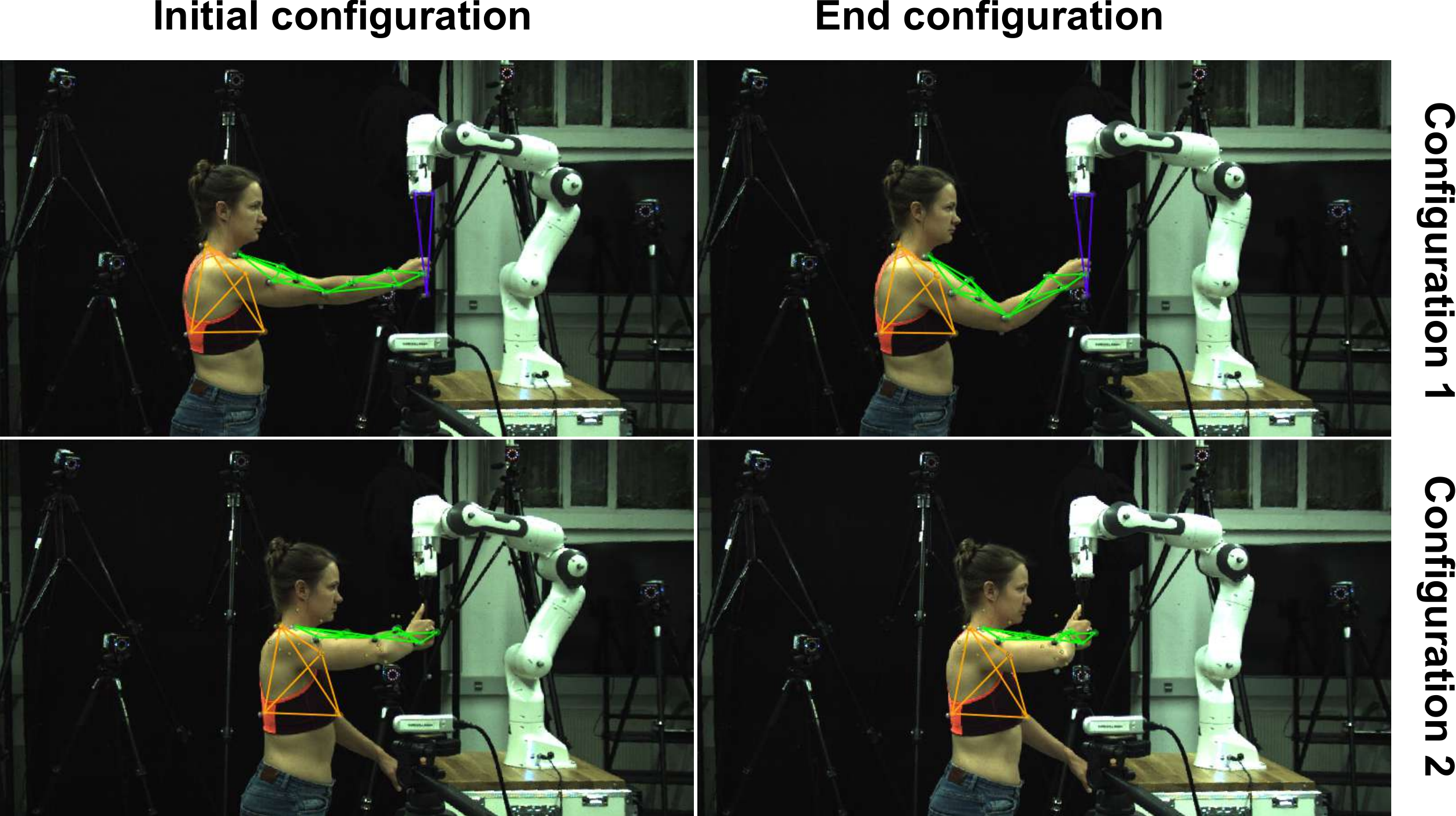}
\caption{Real Experiment setup -- ground truth. The subject moved along the direction index 6; see Fig.~\ref{fig:viconSchem}. The 16 Vicon cameras tracked the attached markers and the body joint angles were continuously measured based on the Vicon built-in upper limb human model.}
\label{fig:viconExp}
\end{figure}

To determine the body Cartesian motions, the participants were instructed to hold and move in sync with a robot end-effector following the specified directions; see Fig~\ref{fig:viconExp}. The Cartesian paths consisted of short straight lines measuring $10$ cm in length, directed towards the hours 3 to 9 on an imaginary horizontal clock face positioned in front of the participants; see Fig.~\ref{fig:viconSchem}. Prior to each motion, the metrics $r$ and $l$ were computed using \eqref{eq:R} and \eqref{eq:l}, respectively, based on the direction $\boldsymbol{\nu}$ and the body configuration $\boldsymbol{q}_h$ recorded by the camera. As elaborated in the following, these metrics can signify the requisite instantaneous joint motions for the corresponding Cartesian motions. To assess their accuracy, the actual resulting body joint motions needed to be determined as the ground truth. To achieve this, a Vicon\textsuperscript{\textregistered} motion capture system equipped with 16 infrared cameras and one HD camera was utilized to record the motion of the right upper limb at $200$ Hz during the experiment. Markers were placed on bony landmarks in accordance with Vicon recommendations for upper limb tracking; see Fig.~\ref{fig:viconExp}. The Vicon Nexus software was employed to compute joint angles using a predefined model template of upper limbs.

Given the length $r$ of the manipulability ellipsoid's radius along the direction $\boldsymbol{\nu}$, the joint velocity magnitude $\|\dot{\boldsymbol{q}}_{h,\boldsymbol{\nu}}\|$ associated with the Cartesian velocity $\dot{\boldsymbol{x}}_{h,\boldsymbol{\nu}}$ along $\boldsymbol{\nu}$ can be estimated using the following expression\footnote{For a unit-norm joint velocity, the Cartesian velocity norm along $\boldsymbol{\nu}$~is~$r$.}.
\begin{align}
\|\dot{\boldsymbol{q}}_{h,\boldsymbol{\nu}}\| = \frac{\|\dot{\boldsymbol{x}}_{h,\boldsymbol{\nu}}\|}{r}. \label{eq:qhnu}
\end{align}
Given that the motion duration along the $10$ cm lines is $1$ second, it can be assumed that $\|\dot{\boldsymbol{x}}_{h,\boldsymbol{\nu}}\| \approx 0.1$ m/s. Utilizing \eqref{eq:qhnu}, the magnitude of the motion in the body joints is estimated according to both the manipulability ellipsoid radius $r$ and the manipulability pseudo-ellipsoid radius $l$ as
\begin{align}
    \delta_r = \frac{0.1}{r}, \hspace{1cm} \delta_l = \frac{0.1}{l}.
\end{align}
The estimated values $\delta_r$ and $\delta_l$ are then compared with the observed joint motions $\|\Delta \boldsymbol{q}_h\|$ recorded by the Vicon system for the three body joints used to derive $r$ and $l$. 

\begin{figure}
\centering
\includegraphics[width=\columnwidth]{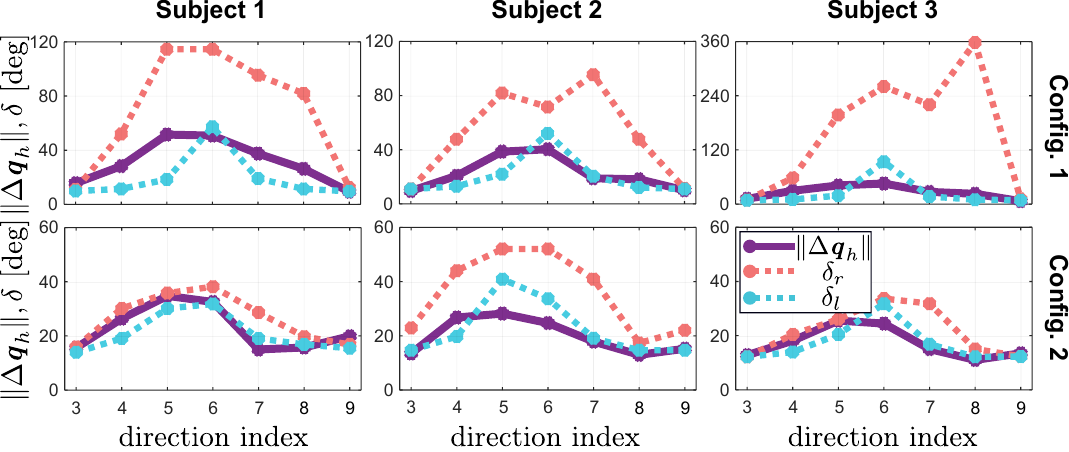}
\caption{Experiment results -- Estimated and actual values of body joints motions for seven Cartesian motions; see Fig.~\ref{fig:viconSchem}. The estimation of the manipulability pseudo-ellipsoid, denoted as $\delta_l$, results in being closer to the actual values of $| \Delta \boldsymbol{q}_h |$ compared to the standard method.}
\label{fig:viconRes}
\end{figure}

As shown in Fig.~\ref{fig:viconRes}, the joint motion estimation $\delta_l$ obtained through the manipulability pseudo-ellipsoid closely aligns with the actual values when compared to the estimated values $\delta_r$ derived from the conventional method. Specifically, during configuration 1, the mean absolute errors between $\delta_l$ and $\|\Delta \boldsymbol{q}_h\|$ are $14$, $7$, and $17$ degrees for subjects 1, 2, and 3, respectively. In configuration 2, the errors are $3$, $5$, and $3$ degrees for the same subjects. In contrast, the mean absolute errors between $\delta_r$ and $\|\Delta \boldsymbol{q}_h\|$ during configuration 1 are $39$, $30$, and $134$ degrees, and during configuration 2 are $4$, $16$, and $5$ degrees for subjects 1, 2, and 3, respectively. The disparity between $\delta_l$ and $\delta_r$ is much more pronounced in configuration 1, where the arm is in close proximity to a singularity, resulting in a much narrower manipulability ellipsoid compared to configuration 2. As mentioned before, a potential explanation for the relatively subpar performance of the conventional manipulability ellipsoid method may be its sensitivity to measurement of body configuration. The Vicon motion capture system boasts significantly higher measurement accuracy compared to the markerless single-camera vision system utilized for manipulability assessment. Thus, especially when sensitivity is heightened, the outcomes could diverge considerably from one another.

Another contributing factor to the significant disparity between $\delta_r$ and $\|\Delta \boldsymbol{q}_h\|$ may be the overly simplified model employed to construct the Jacobian matrix. As previously described, only three joints are taken into consideration in the human arm kinematics model, with the assumption that the other joints remained fixed during the motions, as depicted in Fig.~\ref{fig:kin}. Consequently, the subjects were instructed to focus solely on manipulating the specified three joints during the experiments. However, based on the results from the Vicon system, they inadvertently engaged other joints to a small extent. The average values of $\|\Delta \boldsymbol{q}_h\|$ across all seven directions, when only the primary three joints are considered, are $31$, $22$, and $27$ degrees during configuration 1, and $23$, $20$, and $17$ degrees during configuration 2, for subjects 1, 2, and 3, respectively, as shown in Fig.~\ref{fig:viconRes}. Yet, when six joints (i.e., joint 2 to joint 7, as indicated in Fig.~\ref{fig:kin}) are taken into account, the average values of $\|\Delta \boldsymbol{q}_h\|$ shift to $45$, $28$, and $32$ degrees in configuration 1, and $28$, $22$, and $20$ degrees in configuration 2, for subjects 1, 2, and 3, respectively. The inclusion of these additional joints may have compensated for the need for substantial motions in the primary three joints, which are estimated by the simplified 3-DoF model. Therefore, it is possible that if the additional joints were kept stationary or a more sophisticated kinematics model was utilized, the estimations might have been more accurate. Yet, interestingly, even with this limitation, the estimated values $\delta_l$ are, on average, closer to the actual values $\|\Delta \boldsymbol{q}_h\|$ that included all six joints, compared to the estimated values $\delta_r$. In summary, for the proposed manipulability assessment objective, considering both sensory errors and model inaccuracies, the approach relying on the manipulability pseudo-ellipsoid yields more dependable estimations compared to the conventional method.

\section{Conclusion}
Manipulability analysis evaluates an articulated system's performance in manipulation at each configuration state. Over the past four decades, several approaches have been proposed for this purpose. They share a common aspect: manipulability can be visualized using a pair of dual ellipsoids, each with core matrices inversely related to the other---one associated with motion characteristics and the other with force. Regardless of the chosen approach for constructing the core matrix of the ellipsoid, any inaccuracies in the system configuration or the utilized model can alter the configuration of the ellipsoid. While such inaccuracies may not significantly impact approaches where the volume of the ellipsoid serves as the metric, they can greatly affect direction-dependent manipulability assessment methods, especially when the system is close to singularity. This may not pose an issue for a system like a mechanical arm, where there is direct access to accurate models and sensory information. However, for systems such as the human body, where there is no straightforward access to such information and only estimations can be made, this could jeopardize the manipulability assessment.

To address this challenge, this article extends the standard manipulability analysis approach to develop a method less sensitive to configuration estimation inaccuracies. While retaining the same core matrix, the approach involves projecting the ellipsoid's principal radii onto the intended direction, rather than considering its radius directly along that direction. In practice, this process resembles determining the radius of a newly defined geometric shape—referred to as the manipulability pseudo-ellipsoid—along the intended direction. The pseudo-ellipsoid exhibits fundamental similarities to the original ellipsoid. It shares the same principal radii, and its longest and shortest radii fall within the range of the ellipsoid. Yet, the pseudo-ellipsoid features a smoother surface, which reduces sensitivity to minor changes in its configuration. This claim was substantiated in the article through a sensitivity analysis, followed by a series of simulations and experiments.

Employing the pseudo-ellipsoid for manipulability assessment offers notable advantages, especially in systems like the human arm, where body configuration estimation is necessary. The approach's diminished sensitivity helps to mitigate abrupt changes in assessment caused by data inaccuracies or noise. Should an adaptive controller utilize the manipulability evaluation output, this heightened robustness can effectively prevent sudden shifts in control behavior.

\section*{ACKNOWLEDGMENT}
The authors express sincere gratitude to Petr Svarny for his support with the OpenPose software. This work was funded by the Lighthouse Initiative Geriatronics from StMWi Bayern (Project X, no. 5140951), the BMBF program "Souverän. Digital. Vernetzt.," and the ReconCycle project (no. 871352). M.H. was supported by the European Union under the project ROBOPROX (reg. no. CZ.02.01.01/00/22\_008/0004590). 

\balance
\bibliographystyle{IEEEtran}
\renewcommand{\baselinestretch}{1}\normalsize
\bibliography{library.bib}
\end{document}